\documentclass{article}

\usepackage{PRIMEarxiv}

\usepackage[utf8]{inputenc} 
\usepackage[T1]{fontenc}    
\usepackage{hyperref}       
\usepackage{url}            
\usepackage{booktabs}       
\usepackage{amsfonts}       
\usepackage{nicefrac}       
\usepackage{microtype}      
\usepackage{lipsum}
\usepackage{fancyhdr}       
\usepackage{graphicx}       
\usepackage{amssymb}
\usepackage{amsmath}
\usepackage{latexsym}
\usepackage{amsthm}
\usepackage{booktabs}
\usepackage{enumitem}
\usepackage{graphicx}
\usepackage{color}
\usepackage{tikz}
\usepackage{tabularx}
\usepackage{multirow}
\usepackage{array}
\usepackage[svgnames]{xcolor}
\usepackage{comment}
\usepackage[utf8]{inputenc}
\usepackage{makecell}
\usepackage{url}
\usepackage{todonotes}

\newcommand{\BibTeX}{B\kern-.05em{\sc i\kern-.025em b}\kern-.08em\TeX}
\newcommand{\modelname}{KARAMEL} 
\newcommand{\graphname}{$\mathcal{W}$\textit{iki}$\mathcal{M}$\textit{u}$\mathcal{SA}$}

\newcommand{\artgraph}{$\mathcal{A}rt\mathcal{G}raph$}
\newcolumntype{H}{>{\setbox0=\hbox\bgroup}c<{\egroup}@{}}

\makeatletter
\newcommand{\includeentityimage}[2][\@nil]{%
{ \includegraphics[width=1cm]{#2}}}

\newcommand{\includecandidate}[3][White]{%
\begin{minipage}[t][][t]{\linewidth}
    \colorbox{#1}{%
        \begin{tabularx}{\textwidth}[t]{@{}c@{\hspace{3pt}}>{\raggedright\arraybackslash}X@{}}
            \raisebox{-0.8cm}{\includeentityimage{#2}} & \scriptsize{#3} \\
        \end{tabularx}%
    }
\end{minipage}
}

\makeatother

\definecolor{entcolor1}{named}{Lavender}
\definecolor{entcolor2}{named}{MistyRose}
\colorlet{entcolor3}{Orange!40}
\colorlet{entcolor4}{Lime!40}

\usepackage{subfiles} 

\usepackage{pifont}

\newcommand{\cmark}{\ding{51}} 
\newcommand{\xmark}{\ding{55}} 

\graphicspath{{media/}}     

\title{KARAMEL: Knowledge-Aware Ranking for Multimodal Entity Linking in the Arts via Heterogeneous Graphs
\thanks{\textit{\underline{Citation}}: 
\textbf{Scaringi, R., Sierra-Múnera, A., Vessio, G., Castellano, G., \& Krestel, R.. KARAMEL: Knowledge-Aware Ranking for Multimodal Entity Linking in the Arts via Heterogeneous Graphs (2026). Machine Learning Journal DOI: 10.1007/s10994-026-07172-1.}} 
}

\author{
    Raffaele Scaringi \\
    Department of Computer Science \\
    University of Bari Aldo Moro \\
    Italy \\
    \texttt{raffaele.scaringi@uniba.it}
     \And
    Alejandro Sierra-Múnera \\
    Hasso Plattner Institute \\
    University of Potsdam \\
    Germany \\
     \And
    Gennaro Vessio \\
    Department of Computer Science \\
    University of Bari Aldo Moro \\
    Italy \\ 
     \And
    Giovanna Castellano \\
    Department of Computer Science \\
    University of Bari Aldo Moro \\
    Italy \\
     \And
    Ralf Krestel \\
    ZBW - Leibniz Information Centre for Economics \\
    University of Kiel \\
    Germany \\
}

\begin{document}
\maketitle

\begin{abstract}
Discovering and disambiguating entities from multimodal and heterogeneous data is a key challenge in discovery science, where relevant knowledge is often implicit and distributed across modalities. We present KARAMEL, a framework for multimodal entity linking (MEL)—the task of aligning visual and textual mentions to entities in a knowledge base—in the cultural heritage domain. Unlike prior vision--language models that primarily rely on surface-level multimodal alignment, our approach incorporates sociohistorical and contextual knowledge through a knowledge-aware architecture that fuses a vision--language encoder with a heterogeneous graph neural network. This design enables contextual reasoning over complex relational structures and supports robust entity disambiguation under ambiguous or underspecified evidence. To benchmark multimodal discovery in this setting, we introduce $\mathcal{W}$\textit{iki}$\mathcal{M}$\textit{u}$\mathcal{SA}$ ($\mathcal{W}$ikidata-based $\mathcal{M}$ultimodal $\mathcal{S}$emantic data for $\mathcal{A}$rt), a new dataset that links artworks to Wikidata entities via images, textual descriptions, and structured contextual knowledge. Experimental results show that KARAMEL surpasses strong baselines, particularly on implicit or abstract mentions, while ablation studies highlight the critical role of domain-specific, structured context in learning expressive representations. While instantiated in the cultural heritage domain, KARAMEL addresses a general knowledge discovery problem and applies to other scientific settings involving multimodal observations and heterogeneous background knowledge. \textbf{Both the dataset and source code will be made publicly available upon acceptance.}
\end{abstract}


\section{Introduction}\label{sec1}

In recent years, computer vision and natural language processing have made significant strides, primarily driven by deep learning and large annotated datasets. In many real-world settings, however, the challenge is not only to process visual and textual signals, but to \emph{discover and disambiguate semantic entities} from multimodal and heterogeneous data, where relevant knowledge is often implicit and distributed across modalities. Vision and text transformers~\cite{dosovitskiy2020image, vaswani2017attention} have achieved impressive results in various vision--language tasks, including image--text matching~\cite{jiang2021review}, visual question answering~\cite{ishmam2024image}, image captioning~\cite{stefanini2022show, lu2022data}, and entity linking~\cite{shen2021entity, gjorgjevikj2025foodsem, wang2024entity}. However, these models often process visual and textual modalities in isolation, overlooking the broader contextual information that is crucial for knowledge discovery and robust semantic understanding.

This limitation becomes especially pronounced in the domain of cultural heritage, particularly in the analysis of artworks. Discovering connections among artistic entities requires going beyond digitized artwork images alone. Textual descriptions and structured knowledge bases encode rich social, historical, and political context that is essential for uncovering semantic relationships among entities and supporting reliable link discovery. Vision--text architectures often struggle to capture this contextual knowledge, which is essential for accurate interpretation. Furthermore, visual diversity in art—due to abstraction, stylistic variation, and subjective representation—complicates the use of visual features for tasks such as entity linking~\cite{sierramunera_et_al:TGDK.2.2.8}.

To address these challenges, recent approaches have explored integrating visual data with structured knowledge from knowledge graphs (KGs), achieving promising results in tasks such as artwork classification~\cite{scaringi2025graphclip}. In this context, \textit{multimodal entity linking} (MEL)—the task of identifying and linking entities mentioned in text and depicted in images to a knowledge base—can be viewed as a knowledge discovery problem over complex, multimodal data. Some MEL models represent context by verbalizing entity attributes derived from KGs~\cite{luo2023multi}, but this approach captures only a partial view of an entity’s background. General-purpose KGs such as Wikidata contain extensive cultural heritage information~\cite{Baroncini2023IsWeb}; however, directly applying graph embedding models to such large graphs raises scalability and computational challenges. Recent work in knowledge extraction~\cite{jarnac2023relevant} has therefore focused on identifying relevant subgraphs, though aggressive pruning may introduce inconsistency or data drift, ultimately limiting the reliability of contextual information.

Addressing these challenges requires models that go beyond surface-level multimodal alignment and are able to reason over contextual knowledge. In this work, we propose \modelname{}, a knowledge-aware architecture for multimodal entity linking that explicitly integrates visual and textual information with a third modality: structured contextual knowledge encoded in a heterogeneous knowledge graph. By combining a vision--language backbone with an inductive heterogeneous graph neural network, \modelname{} enables contextual reasoning over relational structures and supports more expressive and accurate entity discovery. We extract a focused artistic subgraph from Wikidata, forming \graphname{}, a lightweight KG tailored to the domain of art, and demonstrate that domain-specific contextual knowledge significantly improves disambiguation beyond vision--text inputs alone.

\begin{figure}[t]
    \centering
    \includegraphics[width=\linewidth]{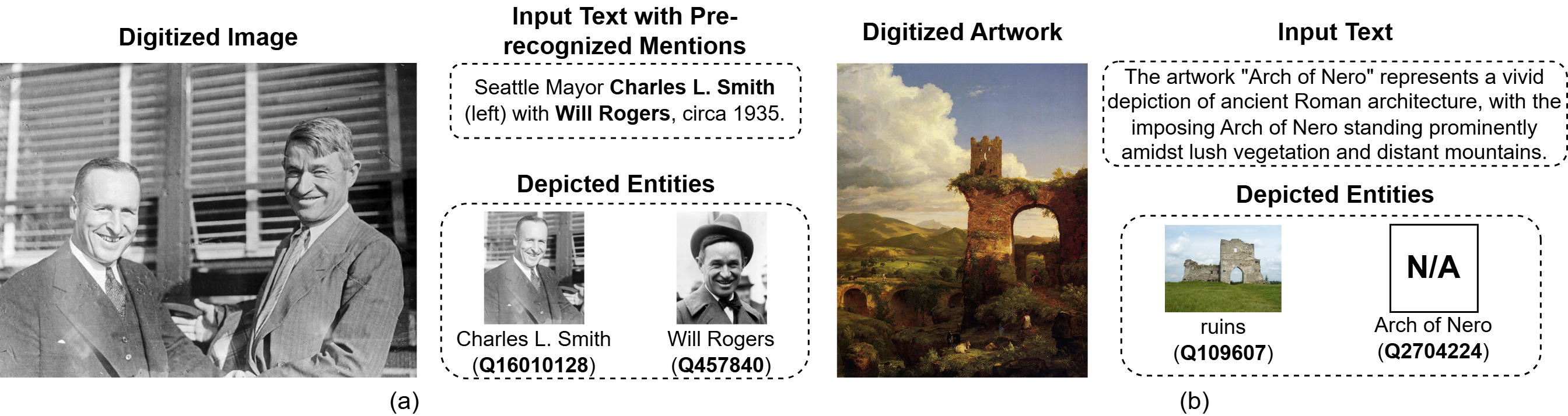}
    \caption{Comparison between the standard multimodal entity linking setting (a), where entity mentions are explicitly highlighted in the text, and our proposed setting (b), where mentions must be inferred from raw Wikipedia articles without predefined annotations.}
    \label{fig:mot-example}
\end{figure}

\modelname{} is specifically designed to address the unique challenges of MEL in the artistic domain. As shown in Fig.~\ref{fig:mot-example}(b), the task involves linking depicted entities from an artwork and its textual description without assuming explicit mentions in the text. For instance, while \textit{Arch of Nero} is mentioned, the link with the entity \textit{ruins} must be inferred from visual evidence combined with contextual cues such as ``\textit{ancient Roman architecture}''. This contrasts with standard MEL settings (Fig.~\ref{fig:mot-example}(a)), where entities are explicitly named and easily extracted. Our dataset, therefore, introduces a more challenging and realistic discovery setting compared to existing benchmarks such as Wikidata-MEL~\cite{wang2022multimodal} or MELArt~\cite{sierramunera_et_al:TGDK.2.2.8}.

A second challenge in this domain is the visual complexity of art: depicted entities may occupy only small image regions, and differences in style or brushwork further complicate visual matching. This further motivates the integration of structured contextual knowledge to support entity discovery and disambiguation beyond what vision--language models can capture in isolation.

The remainder of the paper is organized as follows. Section~\ref{sec:rel_works} reviews related work. Section~\ref{sec:dataset} describes the proposed dataset. Section~\ref{sec:method} introduces our framework. Section~\ref{sec:experiments} presents the experimental evaluation. Finally, Section~\ref{sec:conclusion} concludes the paper and outlines future directions.

\section{Related Work}\label{sec:rel_works}

Entity linking (EL) is the task of mapping textual mentions to corresponding entities within a knowledge base. It can be viewed as a knowledge discovery problem in which latent semantic references must be inferred from incomplete or ambiguous evidence~\cite{dost2020vt, 9560019}. Early methods used word embeddings, such as Word2Vec~\cite {church2017word2vec}, in combination with convolutional neural networks~\cite{li2021survey} to link mentions to entities~\cite{cao2017bridge}. Recognizing the limitations of pure text-based representations, Gupta et al.~\cite{gupta2017entity} proposed incorporating coarse- and fine-grained features to construct more accurate entity representations. 

With the rise of transformer-based architectures~\cite{vaswani2017attention, zhang2024survey}, several works have adapted them to enhance EL. Peters et al.~\cite{peters2019knowledge} introduced a knowledge-enhanced BERT model that jointly encodes mentions and entities using a cross-attention module to produce context-aware representations. Wu et al.~\cite{wu2019scalable} proposed BLINK, a two-stage approach in which BERT encodes mentions and a separate transformer ranks candidate entities. Li et al.~\cite{Li2022EmbracingAI} proposed to inject entity labels as synsets into BERT using an entity-aware adapter. More recently, Zhang et al.~\cite{zhang2024tiger} addressed temporal entity linking by integrating temporal graph data. However, their approach primarily focused on structural and textual signals, without leveraging node-level feature engineering to enhance semantic expressiveness further. 

Although textual signals have proven effective in many EL scenarios, visual information can be critical in disambiguation, especially when entities or mentions are visually grounded. Multimodal entity linking extends traditional EL by incorporating visual features along with textual features~\cite{Bielefeld2024Wiki-VEL:Commons, gan2021multimodal}. Early work by Moon et al.~\cite{moon2018multimodal} showed that combining visual and textual information can alleviate polysemy in social media analysis~\cite{balaji2021machine}. Later studies explored deeper modality interactions: for example, Wang et al.~\cite{wang2022multimodal} used cross-attention mechanisms to fuse visual and textual signals, while Luo et al.~\cite{luo2023multi} introduced a multi-grained matching module to align multimodal entity-mention representations more effectively. 

However, MEL becomes particularly challenging in domains where context plays a central role. The cultural heritage domain exemplifies this, as it involves complex, often implicit, contextual relationships. In recent years, fine arts have become a growing focus within multimodal AI research, leading to advances in tasks such as artwork classification~\cite{scaringi2025graphclip}, style transfer~\cite{guo2024artwork, Dong2025AttenStylerTS}, inpainting~\cite{fanelli2024dream}, and guided artwork generation~\cite{Tang2025ArtCS}. These efforts span predictive and generative settings and have spurred the creation of domain-specific datasets. 

Among them, \artgraph~\cite{castellano2022leveraging} is a large-scale KG constructed from sources such as WikiArt~\cite{castellano2021deep}, DBpedia, and Artemis~\cite{achlioptas2021artemis}, representing artworks, artists, and emotions evoked by artworks. Although rich in general contextual information, \artgraph{} lacks a detailed taxonomy for the depicted entities, limiting its application in fine-grained MEL tasks. Similarly, Artpedia~\cite{conf/iciap/artpedia/19/Stefanini} classifies sentences from Wikipedia painting articles into visual and contextual categories, providing a textual resource for multimodal analysis. Based on this, MELArt~\cite{sierramunera_et_al:TGDK.2.2.8} connects the iconographic entities mentioned in Artpedia’s visual sentences to KG entries, forming an automatically annotated MEL dataset. However, MELArt assumes that the depicted entities are explicitly mentioned in the text and does not aim to cover all entities depicted in an artwork, thereby limiting its generalizability. 

To address these gaps, we propose a knowledge-aware perspective on multimodal entity linking that explicitly leverages structured relational context for discovery and disambiguation. Within this framework, we introduce \graphname{}, a curated subgraph extracted from Wikidata that focuses on domain-relevant relationships in the artistic domain, balancing information richness and scalability. Building on this resource, we present \modelname{}, a multimodal framework that integrates images, text, and heterogeneous graph-based knowledge to learn expressive representations of depicted entities.

\section{\graphname{}}\label{sec:dataset}

Understanding artworks and the entities they depict requires discovering semantic relationships across multimodal, heterogeneous sources, where relevant context is often implicit rather than explicitly observed. Visual and textual cues alone are frequently insufficient, as historical, social, political, or religious background plays a central role in interpretation. For example, knowing that another mentored an artist or that a depicted figure held a significant public role or personal relationship can enrich interpretation. This contextual layer can be modeled through both structured metadata and unstructured descriptions. A comprehensive knowledge graph allows such connections to be explicitly represented, capturing interdependencies and domain semantics.

\begin{figure}[!t]
    \centering
    \includegraphics[width=\textwidth]{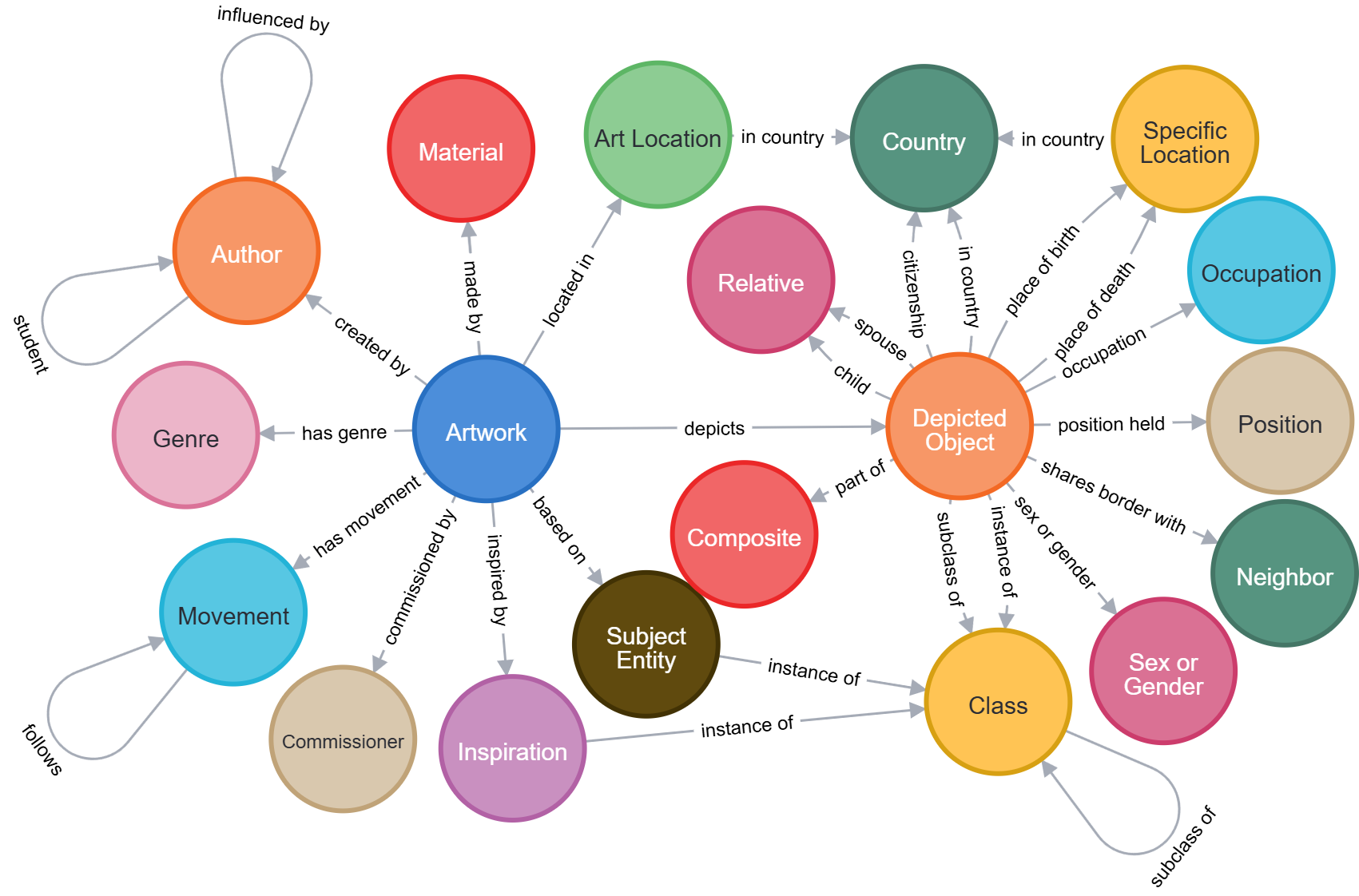}
    \caption{Logical schema of \graphname{}. Each artwork is associated with metadata, including author, genre, and movement. Depicted entities are linked through enriched, contextual relationships that enhance expressiveness and graph connectivity.}
    \label{fig:graph}
\end{figure}

To this end, we introduce \graphname{}, a dataset in the artistic domain comprising visual, textual, and contextual information relevant to cultural heritage. Figure~\ref{fig:graph} illustrates its core schema, focusing on the relationships of entities and their domain-specific properties.

\subsection{Data Collection and Processing}

We constructed \graphname{} from artworks in Wikidata that meet the following criteria: ($i$) a linked \textit{Wikipedia article}; ($ii$) at least one \textit{depicts} relation to another Wikidata entity; and ($iii$) an associated \textit{image}. Wikipedia articles were retrieved using the Huggingface dataset \textit{wikimedia/wikipedia},\footnote{\url{https://huggingface.co/datasets/wikimedia/wikipedia}} based on the 2023-11-01 dump. Since articles span multiple languages, we used the NLLB-200 (3.3B) model~\cite{DBLP:journals/corr/abs-2207-04672} to translate non-English texts into English. We used LLaMA 3 (8B)~\cite{grattafiori2024llama3herdmodels} to generate depiction-focused summaries, thereby limiting article length and reducing variance. This summarization step was designed to reduce variance and noise across heterogeneous sources while preserving depiction-relevant content, and was applied consistently across all models to ensure a fair and controlled experimental setting. The distribution of original languages and the length of articles before and after summarization are discussed in detail in Section~\ref{sec:ap_translation_summarization}.

\subsection{Graph Structure and Relations}

Each artwork is represented as a node connected to multiple nodes of contextual information, including author, genre, artistic movement, gallery, and depicted entities. These depicted entities can be people, places, abstract concepts, and more. For each category, we selected a curated set of relevant properties: \textit{people} (occupation, gender, relatives); \textit{places} (neighboring cities and location hierarchy, e.g., a city within a country), and \textit{galleries} linked to the city and hosting relationships. Each node in \graphname{} could include visual and textual features. Visual data, if available, is sourced from the Wikidata image property (linked to Wikimedia Commons), while textual information is created by concatenating the entity's English label and description. 

\subsection{Final Dataset}

The resulting dataset is a multimodal resource (vision, language, context) comprising $13,482$ unique artworks and $12,177$ unique depicted entities. Each artwork is associated with an image and a summarized Wikipedia article, while each depicted entity is associated with an optional image, a description, and a structured context from \graphname{}. Hence, artworks are modeled using image-text pairs, with the depicted entities represented by visual features, textual descriptions, and graph-derived contextual embeddings. This setup provides a rich testbed for multimodal entity linking in the arts and, more broadly, for studying knowledge-aware discovery from multimodal and heterogeneous data in domains where contextual information plays a central role.

\section{Proposed Method}\label{sec:method}

We propose \modelname{}, a context-aware framework for multimodal entity linking that explicitly addresses entity discovery under ambiguous and incomplete evidence. Building on recent advances in vision–language models, \modelname{} extends vision–text alignment by incorporating structured contextual knowledge from a heterogeneous knowledge graph through a dedicated graph encoder.

\subsection{Problem Formulation}\label{sec:problem_formulation}

The goal of MEL is to link a set of multimodal \textit{mentions} to corresponding \textit{entities} using multiple information sources, such as visual and textual signals. Formally, we define the mention set as $\mathcal{M} = \{\mathbf{M}_i\}_{i=1}^N$, where each mention $\mathbf{M}_i$ consists of a visual representation $M_{vi}$ and a textual description $M_{ti}$, that is, $\mathbf{M}_i = (M_{vi}, M_{ti})$. Similarly, the set of candidate entities is $\mathcal{E} = \{\mathbf{E}_j\}_{j=1}^P$, with each entity $\mathbf{E}_j = (E_{vj}, E_{tj})$ including an image and a textual description.

We also introduce a heterogeneous knowledge graph $\mathcal{G} = (\mathcal{V}, \mathcal{L})$, where $\mathcal{V} = \{v_1, \dots, v_{\vert\mathcal{V}\vert}\}$ is the set of nodes and $\mathcal{L} \subseteq (\mathcal{V}\times \mathcal{V}) = \{l_1, \dots, l_{\vert\mathcal{L}\vert}\}$ is the set of edges. Each node $v \in \mathcal{V}$ is associated with a type via a function $\mu \colon \mathcal{V} \rightarrow \mathcal{T}_{\mathcal{V}}$, where $t_v = \mu(v)$ denotes its node type (e.g., \textit{Artwork}, \textit{Depicted Object}, \textit{Artist}). Additionally, we define the support of each node type $ t$ as the set of nodes associated with $t$. In general, $\mathcal{V}_{t} = \{v \in \mathcal{V} \ \vert \mu(v) = t\}$. Each edge $l \in \mathcal{L}$ is similarly typed through a function $\rho \colon \mathcal{L} \rightarrow \mathcal{T}_{\mathcal{L}}$, where $t_l = \rho(l)$ defines its semantic relation (e.g., \textit{depicts}, \textit{located in}, \textit{instance of}, etc.). We assume that for every entity $\mathbf{E}_j \in \mathcal{E}$ there exists a corresponding node $v_j \in \mathcal{V}$ of type \textit{Depicted Object}. Given an input mention $\mathbf{M}_i$, the goal of MEL is to correctly identify the ground-truth entity $\mathbf{E}_j \in \mathcal{E}$ by maximizing its probability among all candidates. This task can be formalized as the maximization of the log-likelihood over a training set $\mathcal{D}_{\text{train}}$:
\begin{equation} \label{eq:base_mel} \theta^* = \arg\max_{\theta} \sum_{(\mathbf{M}_i, \mathbf{E}_j) \in \mathcal{D}_{\text{train}}} \log p_{\theta}(\mathbf{E}_j \mid \mathbf{M}_i, \mathcal{E}). \end{equation}
To incorporate contextual knowledge from the KG, we extend the objective to condition on $\mathcal{G}$ as well:
\begin{equation} \theta^* = \arg\max_{\theta} \sum_{(\mathbf{M}_i, \mathbf{E}_j) \in \mathcal{D}_{\text{train}}} \log p_{\theta}(\mathbf{E}_j \mid \mathbf{M}_i, \mathcal{E}, \mathcal{G}). \end{equation}
This formulation encourages the model to leverage all three modalities—vision, text, and context—during entity ranking.

\subsection{Model Architecture} \label{sec:model-arc}

\begin{figure}[!t]
    \centering
    \includegraphics[width=\linewidth]{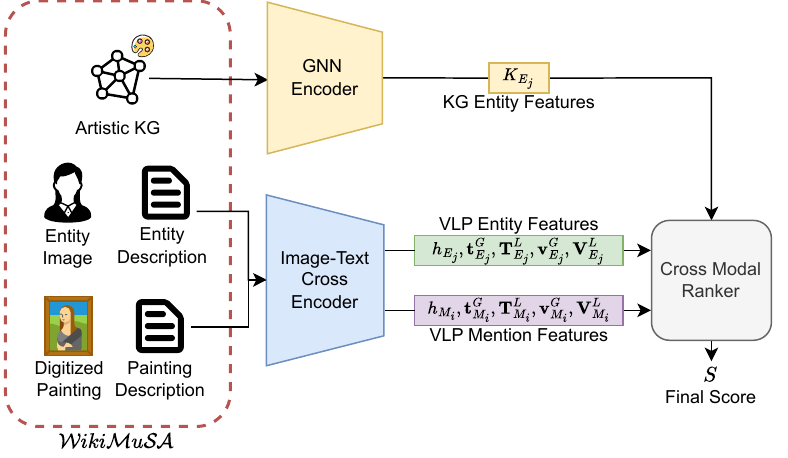}
    \caption{Architecture of \modelname{}. The image-text cross-encoder and the GNN encoder are jointly trained to extract complementary information from artworks and depicted entities. The cross-modal ranker is then trained to rank candidate entities effectively for a given input mention.}
    \label{fig:framework}
\end{figure}

\modelname{} jointly processes an artwork mention $\mathbf{M}_i$, a candidate entity $\mathbf{E}_j$, and a knowledge graph $\mathcal{G}$ using three main architectural components: an image-text cross-encoder, a heterogeneous graph neural network (GNN) encoder, and a cross-modal ranker, as illustrated in Fig.~\ref{fig:framework}. The image-text cross-encoder extracts expressive features from both images and textual descriptions associated with artwork mentions and depicted entities, using a CLIP architecture~\cite{pmlr-v139-radford21a} pre-trained on the LAION-400M dataset~\cite{schuhmann2021laion}, whereas the GNN encoder exploits graph convolutions and the structured semantics in the input KG to extract context-aware information for all depicted entities, improving their expressiveness.

\subsubsection{Feature Encoding} \label{sec:feat-enc}

\paragraph{Visual Feature Extraction}

The CLIP vision encoder employs a vision transformer (ViT)~\cite{dosovitskiy2020image} as its backbone. Given an RGB image $I$ with dimensions $C \times H \times W$—either the artwork image $M_{vi}$ or the entity image $E_{vj}$—ViT encodes it into a hidden representation $\mathbf{V} \in \mathbb{R}^{(n+1) \times d}$: $\mathbf{V} = \left[\mathbf{v}_{[CLS]}^0, \mathbf{v}^1, \dots, \mathbf{v}^n\right]$. We extract two types of visual features: the global visual representation, corresponding to the [CLS] token, $\mathbf{v}_G \in \mathbb{R}^d$; and the local visual representation, given by all token embeddings, $\mathbf{V}_L \in \mathbb{R}^{(n+1) \times d}$. For artwork mentions, these are denoted as $\mathbf{v}_{M_i}^G$ and $\mathbf{V}_{M_i}^L$, respectively. For entities, we similarly define $\mathbf{v}_{E_j}^G$ and $\mathbf{V}_{E_j}^L$. When entities lack visual information, we apply zero-padding as standard practice.

\paragraph{Textual Feature Extraction}

For textual encoding, we use a pre-trained BERT model~\cite{devlin2019bert}. Given an input text sequence $\mathbf{S}$ consisting of $l_s$ tokens, BERT produces a contextualized embedding matrix $\mathbf{T} \in \mathbb{R}^{(l_s+1) \times d} \colon \mathbf{T} = \left[\mathbf{t}_{[CLS]}^0, \mathbf{t}^1, \dots, \mathbf{t}^{l_s}\right]$. As with visual features, we extract the global textual representation of the [CLS] token, $\mathbf{t}_G \in \mathbb{R}^d$, and the local textual representation of the entire token sequence, $\mathbf{T}_L \in \mathbb{R}^{(l_s + 1) \times d}$. We denote the textual representations of the artwork mentions as $\mathbf{t}_{M_i}^G$ and $\mathbf{T}_{M_i}^L$, and those of entities as $\mathbf{t}_{E_j}^G$ and $\mathbf{T}_{E_j}^L$.

\paragraph{Graph Feature Extraction}\label{par:graph-feat-extraction}

The graph encoder employs a GNN to extract contextual embeddings from $\mathcal{G}$, focusing on nodes representing depicted entities. Let $t = \text{``depicted entity''}$ be the target node type. The consequent knowledge graph embedding matrix encompassing all candidate depicted entities can be defined as: 
\begin{equation}\label{eq:ent_kg_embs}
    \mathbf{K} = \left[\mathbf{k}_{E_j} = \mathbf{x}_{v_j}^{(L)} \ \vert v_j  \in \mathcal{V}_t\right],
\end{equation}
where each depicted entity $E_j \in \mathcal{E}$ is encoded as $\mathbf{k}_{E_j} \in \mathbb{R}^d$, $\mathbf{x}_{v_j}^{(L)}$ is the learned embedding of node $v_j$ at the last GNN layer $L$, and $\mathcal{V}_t$ is the set of nodes having type $t$. 

In general, each GNN layer updates node representations by aggregating information from neighboring nodes through permutation-invariant operations and transforming it via learnable functions. This message-passing mechanism enables nodes to incorporate contextual information from their local graph neighborhood. Node neighborhoods are defined by the graph connectivity, and the aggregation process is applied consistently across all nodes. Initial node features are derived from node-specific metadata; in our implementation, we encode each Wikidata entity by concatenating its label and description, then project the resulting text into a dense embedding space using a pre-trained BERT model.

\subsubsection{Cross-Modal Ranker}\label{sec:crossmodal-ranker}

To compute a ranking score for each mention-entity pair, \modelname{} combines multiple modalities to explore intra- and inter-modal interactions among visual, textual, and knowledge graph information. Specifically, we define four components that contribute to the overall score: a vision-based score, a text-based score, a vision-text cross-modal score, and a contextual knowledge score.

\paragraph{Vision-Based Score}

Visual information is fundamental in MEL, particularly in the art domain, where artworks are primarily expressed through visual characteristics: from low-level features like brushstrokes to high-level representations such as depicted objects and symbolic motifs. We adopt the dual-gated mechanism introduced in MIMIC~\cite{luo2023multi} to model visual interaction, which computes the cross-attention between mention and entity features. Formally, the vision-based score is computed as:
\begin{equation}\label{eq:vscore}
    S_V = \mathcal{U}_V(\mathbf{M}_i, \mathbf{E}_j) = (S_V^{E2M} + S_V^{M2E})/2, \ S_V \in \mathbb{R}
\end{equation}
where $S_V^{E2M}$ and $S_V^{M2E}$ are obtained using cross-visual attention between mention and entity representations, having $S_V^{A2B} = \Phi_{A2B}(\mathbf{v}_A^G, \mathbf{v}_B^G, \mathbf{V}_B^L)$, where $\Phi_{A2B}(\cdot, \cdot, \cdot)$ is the scoring attention module, and $\mathbf{v}_A^G$ and $\mathbf{V}_B^L$ are the global and local vision representations defined in Sec.~\ref{sec:feat-enc}.

\paragraph{Text-Based Score}

Text remains a core modality in MEL. Following~\cite{luo2023multi}, we compute a text-based score using the text-based global-local interaction unit (TGLU), which models both global and local textual dependencies:
\begin{equation}\label{eq:tscore}
    S_T = \mathcal{U}_T(\mathbf{M}_i, \mathbf{E}_j) = (S_T^{G2G} + S_T^{G2L})/2,\ S_T \in \mathbb{R}
\end{equation}
where $S_T^{G2G} \in \mathbb{R}^d$ is the global-to-global score, formalized as $S_T^{G2G} = \mathbf{t}_{E_j}^G \cdot \mathbf{t}_{M_i}^G$, and the global-to-local score is a gated cross-attention mechanism $S_T^{G2L} = \mathbf{t}_{E_j}^G \cdot ATTN(\mathbf{T}_{E_j}^L, \mathbf{T}_{M_i}^L, \mathbf{T}_{M_i}^L), \ S_T^{G2L} \in \mathbb{R}^d$.

\paragraph{Vision-Text Cross-Modal Score}

To jointly model visual and textual information, we compute a vision-text cross-modal score:
\begin{equation}\label{eq:cross-vis-text}
    S_{VT} = \mathcal{U}_{VT}(\mathbf{M}_i, \mathbf{E}_j) =  \mathbf{h}_{E_j}^T \mathbf{h}_{M_i}, \ S_{VT} \in \mathbb{R}
\end{equation}
where $\mathbf{h}_{E_j}\in \mathbb{R}^d$ and $\mathbf{h}_{M_i}\in \mathbb{R}^d$ are cross-modal representations regarding the mention $M_i$ and the entity $E_j$, respectively. More precisely, those cross-attended representations are obtained as $\mathbf{h}_{M_i} = CMFU(\mathbf{V}_{M_i}^L, \mathbf{T}_{M_i}^L, \mathbf{v}_{M_i}^G, \mathbf{t}_{M_i}^G)$, or for an entity $E_j \in \mathcal{E}$ $\mathbf{h}_{E_j} = CMFU(\mathbf{V}_{E_j}^L, \mathbf{T}_{E_j}^L, \mathbf{v}_{E_j}^G, \mathbf{t}_{E_j}^G)$,  where $CMFU$~\cite{luo2023multi} embodies a multi-grained cross-attention strategy for gathering a comprehensive representation of the mention or the entity.

\paragraph{Contextual Knowledge Score}

Finally, we integrate contextual information from the knowledge graph by computing an alignment between the mention’s cross-modal representation and the entity’s KG embedding:
\begin{equation}\label{eq:crossscore}
    S_C = \mathcal{U}_C(\mathbf{M}_i, \mathbf{E}_j) =  \mathbf{k}_{E_j}^T \mathbf{h}_{M_i}, \ S_C \in \mathbb{R}
\end{equation}
where $\mathbf{k}_{E_j}\in \mathbb{R}^d$ is the knowledge graph embedding for the entity $E_j \in \mathcal{E}$ obtained using Eq.~\ref{eq:ent_kg_embs}, and $h_{M_i} \in \mathbb{R}^d$ is the cross-attended vision-text mention representation.

\paragraph{Final Ranking Score}

Combining all the components above, the final relevance score for a mention-entity pair is defined as the average of the four scores:
\begin{equation}\label{eq:score}
    S = \mathcal{U}(\mathbf{M}_i, \mathbf{E}_j) = (S_V + S_T + S_{VT} + S_C) / 4
\end{equation}

This ranking function enables \modelname{} to reason beyond surface features by incorporating structured contextual knowledge, capturing fine information such as historical, religious, and social factors. Unlike current frameworks, our knowledge-aware design facilitates discovery and disambiguation in multimodal settings by leveraging structured relational context. While focused on the artistic domain, the approach is applicable to a range of multimodal scenarios.

\section{Experiments}\label{sec:experiments}

We conducted a comprehensive set of experiments on the proposed dataset to answer the following research questions:
\begin{description}
    \item[RQ1] How does \modelname{} perform in ranking candidate entities compared to the baseline models? 
    \item[RQ2] Which characteristics of the data most influence the performance of \modelname{}? 
    \item[RQ3] How does model performance vary under different raw node feature encoding strategies?
\end{description} 
    
We evaluated \modelname{} against several entity linking baselines, training all models on the proposed dataset \graphname{} (see Sec.~\ref{sec:dataset}). During the evaluation, the candidate universe is the complete set of depicted entities. These are ranked for each artwork in the test set, and we report standard retrieval metrics: recall at rank $k$ (R@K), mean average precision (MAP), and mean reciprocal rank (MRR). In our task, where each input mention may have multiple relevant entities, MAP captures the average precision across all ground-truth entities. At the same time, MRR focuses only on the first correctly ranked entity. For all metrics, higher values are better.

\subsection{Experimental Setting}

We randomly split \graphname{} into training, validation, and test sets using an 80/10/10 ratio, based on artworks. To prevent data leakage, all validation and test artworks were removed from the graph $\mathcal{G}$. This ensures that no validation or test artwork nodes are present in the graph, thereby eliminating any indirect influence on the entity embeddings' learning objective. Furthermore, all images (for mentions and entities) were resized to $224 \times 224$ pixels. For entities without associated images, we applied zero-padding. For visual and textual feature encoding (see Sec.~\ref{sec:feat-enc}), we used a pre-trained CLIP (base) model~\cite{pmlr-v139-radford21a}. For the graph encoder, we trained a GNN from scratch using GraphSAGE layers~\cite{hamilton2017inductive}.

Regarding textual input for mentions, we diverged from the standard entity linking formulation, which typically formats mentions as:
\begin{equation}
    M_{it} = [\text{CLS}] \mathbf{m}_l [\text{SEP}] \mathbf{m} [\text{SEP}] \mathbf{m}_r,
\end{equation}
that is, the mention surrounded by left and right contexts. Instead, we use the full article text without predefining mention spans. This challenges the model to identify relevant text portions associated with each entity, rather than relying on annotated mentions, and, at the same time, allows models to use a single mention representation for multiple depicted entities.

Inspired by Luo et al.~\cite{luo2023multi}, we employed a unit-consistent contrastive loss that accounts for all three modalities. Based on the composite scores defined in Eqs.~\ref{eq:vscore}–\ref{eq:score}, we first defined the main contrastive loss as:
\begin{equation}
    \mathcal{L}_{O} = -\log\frac{\exp(\mathcal{U}(\mathbf{M_i}, \mathbf{E}_j))}{\sum_k{\exp(\mathcal{U}(\mathbf{M}_i, \mathbf{E}_k))}},
\end{equation}
where $\mathbf{E}_k$ denotes negative samples within the same batch. To promote disentangled representations and assess the contribution of individual modalities, we further computed auxiliary losses for each interaction component:
\begin{equation}
    \mathcal{L}_{X} = -\log\frac{\exp(\mathcal{U}_X(\mathbf{M_i}, \mathbf{E}_j))}{\sum_k{\exp(\mathcal{U}_X(\mathbf{M}_i, \mathbf{E}_k))}}, X \in \{T, V, VT, C\}.
\end{equation}
Here, $X$ refers to the textual ($T$), visual ($V$), vision-text ($VT$), and contextual ($C$) interaction scores defined in Sec.~\ref{sec:crossmodal-ranker}. The final loss function is the sum of all components:
\begin{equation}
    \mathcal{L} = \mathcal{L}_O + \mathcal{L}_V + \mathcal{L}_T + \mathcal{L}_{VT} + \mathcal{L}_C.
\end{equation}
This promotes consistency across modalities and ensures robust mention-entity matching.

We trained the model for 150 epochs with a batch size of 64. We used the AdamW optimizer~\cite{loshchilov2017decoupled}, a step-wise cosine decay scheduler~\cite{loshchilov2016sgdr}, and a linear warm-up of 1,670 steps (equal to the number of training batches). The maximum learning rate was set to $1e{-5}$. We applied early stopping based on MAP and terminated training if no improvement was observed for 10 consecutive validation epochs.

All experiments were repeated five times with different seeds. The experiments were conducted on a single NVIDIA A100 GPU (64 GB VRAM) with 512 GB RAM, ensuring sufficient computational resources for training on our large multimodal dataset.

\subsection{Baselines}

We compared \modelname{} against four representative baselines, covering text-only, vision-language pre-trained (VLP), and multimodal models. Specifically, we evaluated: 
\begin{itemize}
    \item BERT~\cite{devlin2019bert}, as a strong unimodal text-based model;
    \item CLIP~\cite{pmlr-v139-radford21a} and ViLT~\cite{kim2021vilt}, as popular VLP models, representing straightforward multimodal baselines;
    \item MIMIC~\cite{luo2023multi}, a state-of-the-art multimodal model tailored for MEL that integrates multigrained analysis of vision and text. 
\end{itemize}

This setup enabled us to evaluate the impact of various modality combinations on the representation of mentions and candidate entities. In particular, we analyzed how relying solely on textual features compares to leveraging joint vision-text representations and how including contextual knowledge further improves performance. 

\subsection{Experimental Results}

\subsubsection{Performance Comparison}

\begin{table*}[!t]
\centering
\caption{Ranking metrics for the baselines and \modelname{}. Each cell reports the mean and standard deviation (in parentheses). All metrics are expressed in the range [0, 100]. The best performance per metric is shown in \textbf{bold}.}
\resizebox{\textwidth}{!}{%
\begin{tabular}{lcHcHcccc}
\toprule
Model & R@1 & R@5 & R@10 & R@30 & R@50 & R@100 & MAP & MRR \\
\midrule
BERT~\cite{devlin2019bert} & 2.33 ($\pm$0.88) & 7.27 ($\pm$2.24) & 10.56 ($\pm$2.53) & 18.97 ($\pm$3.71) & 24.52 ($\pm$3.61) & 32.41 ($\pm$3.2) & 5.93 ($\pm$1.53) & 11.94 ($\pm$2.36) \\
CLIP~\cite{pmlr-v139-radford21a} & 0.24 ($\pm$0.08) & 1.21 ($\pm$0.32) & 2.68 ($\pm$0.68) & 8.08 ($\pm$1.61) & 13.12 ($\pm$2.9) & 22.57 ($\pm$4.12) & 1.25 ($\pm$0.09) & 3.0 ($\pm$0.19) \\
ViLT~\cite{kim2021vilt} & 0.62 ($\pm$0.1) & 2.66 ($\pm$0.33) & 5.01 ($\pm$0.58) & 11.44 ($\pm$0.59) & 16.4 ($\pm$0.51) & 23.71 ($\pm$1.27) & 1.08 ($\pm$0.22) & 2.43 ($\pm$0.5) \\
MIMIC~\cite{luo2023multi} & 27.22 ($\pm$1.19) & 51.82 ($\pm$1.24) & 60.93 ($\pm$1.32) & 72.49 ($\pm$0.95) & 77.26 ($\pm$0.75) & 82.52 ($\pm$0.62) & 47.0 ($\pm$1.66) & 65.72 ($\pm$1.88) \\
\modelname{} (\textit{Ours}) & \textbf{27.26 ($\pm$0.47)} & \textbf{52.82 ($\pm$0.67)} & \textbf{62.31 ($\pm$0.57)} & \textbf{73.75 ($\pm$0.66)} & \textbf{78.48 ($\pm$0.67)} & \textbf{83.65 ($\pm$0.45)} & \textbf{48.45 ($\pm$0.7)} & \textbf{66.88 ($\pm$0.51)} \\ 
\bottomrule
\end{tabular}}
\label{tab:results}
\end{table*}

To answer RQ1, we compare \modelname{} with the selected baselines in Table~\ref{tab:results}. Across all metrics, \modelname{} outperforms the baselines, with MIMIC being the closest competitor. This demonstrates that incorporating contextual information from a knowledge graph leads to more accurate and robust entity linking, particularly in complex domains such as cultural heritage.

Although being a unimodal baseline, BERT achieves surprisingly strong results, outperforming CLIP and ViLT despite using only textual information. This suggests that the modality gap~\cite{liang2022mind} in vision-language pre-training may hinder the performance of contrastive models, such as CLIP and ViLT, in this setting. Moreover, BERT benefits from dense textual supervision, which is especially useful when entities are described in rich natural language, as in Wikipedia.

Compared to MIMIC, \modelname{} yields consistent improvements across all metrics, with a particularly clear gain in MAP and a lower variance across seeds, indicating more stable ranking behavior. Beyond aggregate performance, the benefit of structured context becomes most apparent in discovery-oriented cases where evidence is incomplete—e.g., when entities are not explicitly mentioned in the text or when depictions are more abstract—supporting the role of graph-based contextual reasoning for disambiguation. This also shows that encoding entities using a KG, in this case \graphname{}, is beneficial for exploiting correlations among entities, detected by graph convolutions. For example, knowing that a candidate entity is depicted in specific artworks (i.e., the \textit{depicts} relationship—first-order information) provides insight into the artistic expressiveness associated with that entity. Additionally, leveraging metadata about those artworks (e.g., their \textit{styles}, \textit{genres}, etc.—second-order information) enriches the learned entity representation by highlighting the variations in how an entity may be portrayed depending on these factors. Finally, recognizing that multiple entities, for example, share the same social position (i.e., the \textit{position held} relationship—first-order information) enables the model to exploit these correlations, resulting in more robust representations. In contrast, MIMIC relies solely on vision-text information and is unable to exploit the aforementioned features, resulting in a pure independent-entity encoding strategy that performs worse.

\begin{table*}[!t]
\centering
\caption{Ranking metrics for MIMIC and \modelname{} on MELArt~\cite{sierramunera_et_al:TGDK.2.2.8}. Each cell reports the mean and standard deviation (in parentheses). All metrics are expressed in the range [0, 100]. The best performance per metric is shown in \textbf{bold}.}
\resizebox{\textwidth}{!}{%
\begin{tabular}{lcHcHcccc}
\toprule
Model & R@1 & R@5 & R@10 & R@30 & R@50 & R@100 & MAP & MRR \\
\midrule
MIMIC~\cite{luo2023multi} & 50.07 ($\pm$1.98) & 78.07 ($\pm$2.16) & \textbf{83.0 ($\pm$1.47)} & \textbf{89.36 ($\pm$0.62)} & \textbf{91.76 ($\pm$1.28)} & \textbf{94.73 ($\pm$0.91)} & 70.08 ($\pm$1.96) & 80.44 ($\pm$1.67) \\
\modelname{} & \textbf{51.07 ($\pm$1.62)} & \textbf{78.74 ($\pm$1.65)} & 82.91 ($\pm$2.35) & 88.79 ($\pm$2.08) & 91.09 ($\pm$2.09) & 94.25 ($\pm$1.7) & \textbf{71.34 ($\pm$1.92)} & \textbf{81.34 ($\pm$1.57)} \\
\bottomrule
\end{tabular}}
\label{tab:results_melart}
\end{table*}

We further evaluate \modelname{}'s performance against MIMIC using MELArt~\cite{sierramunera_et_al:TGDK.2.2.8} for training and evaluation. We adapt the dataset to our experimental setting by extracting a graph with the same ontology as \graphname{} seen in Figure~\ref{fig:graph}, containing the artworks and depicted entities in MELArt. Furthermore, since MELArt is annotated at the entity-mention level, with artwork texts separated into independent sentences, we concatenate all sentences from the same artwork into a longer textual description. We constrain the candidate set to the depicted entities. The results shown in Table~\ref{tab:results_melart}, show that \modelname{} improves over MIMIC in retrieving relevant depicted entities in the first position of the ranking and in general shows better ranking positions for the known depictions as seen by MAP and MRR values. However there is degradation of recall in the rest of the recall thresholds. In general the difference between the models is smaller and the standard deviation between runs is higher. We hypothesize that this is due to MELArt being a considerably smaller dataset than~\graphname{}.

\subsubsection{Feature-Wise Statistical Analysis}
\begin{figure}[!t]
    \centering
    \includegraphics[width=\textwidth]{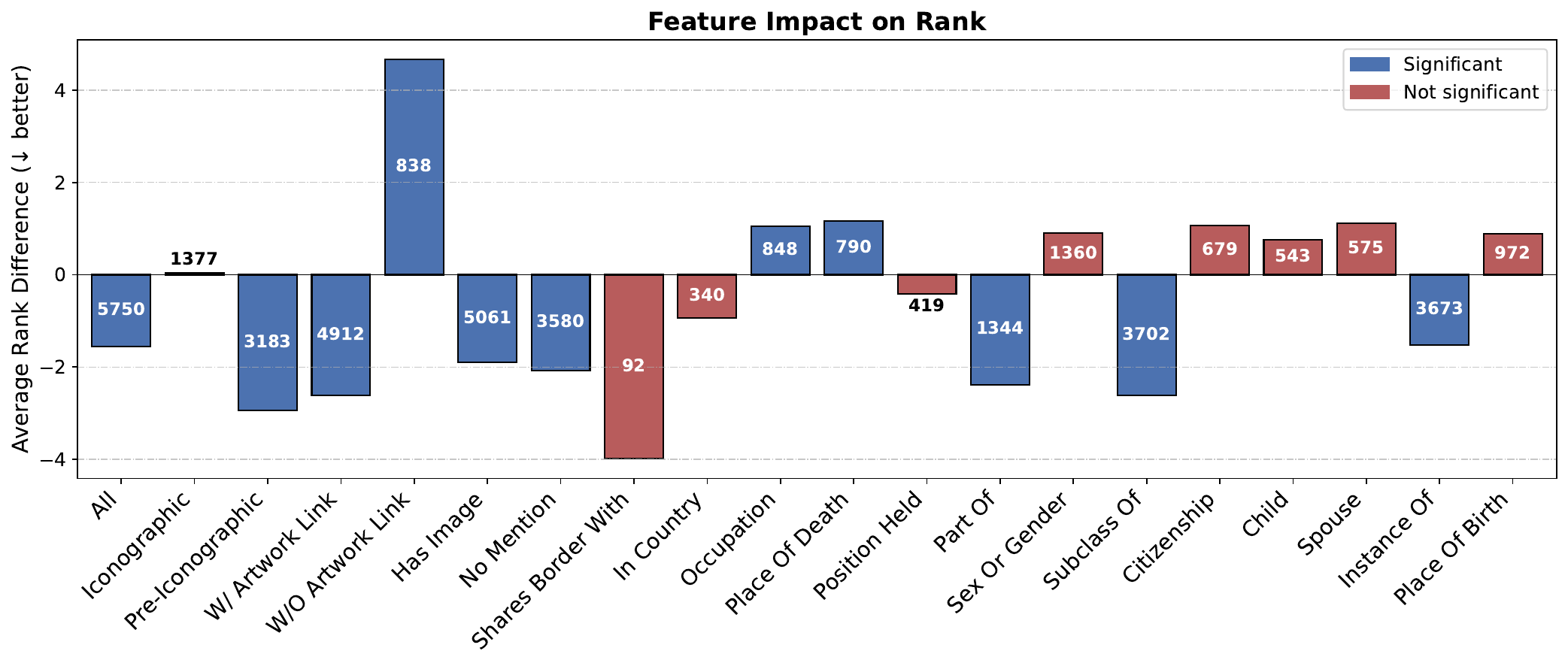}
    \caption{Average rank difference of \modelname{} on MIMIC~\cite{luo2023multi}. Bars below the x-axis indicate improved ranking of \modelname{}. Numbers within each bar denote the number of (\textit{artwork}, \textit{depicted entity}) pairs supporting the corresponding feature. Blue bars denote statistically significant effects, whereas red bars indicate non-significant effects.}
    \label{fig:feat_importance}
\end{figure}

To address RQ2 and validate the effectiveness of the ontological information in \graphname{}, Fig.~\ref{fig:feat_importance} presents a detailed statistical significance analysis comparing the rankings produced by \modelname{} and MIMIC. Specifically, we considered all (\textit{artwork}, \textit{depicted entity}) test pairs and computed the rank difference as the rank predicted by \modelname{} minus that predicted by MIMIC. We applied both a one-sided Wilcoxon signed-rank test and a paired t-test to assess the statistical significance of these differences. A difference was deemed statistically significant only if both tests yielded $p$-values below 0.05. Throughout our analysis, a negative value indicates that \modelname{} ranked the correct entity higher than MIMIC (i.e., closer to the top), reflecting improved performance, whereas a positive value indicates worse performance. As a base case, if the ground-truth entity did not appear in the top 100 results for either model, we assigned it a default rank of 101.

We further enriched our analysis by reporting quantitative results based on both ontological and non-ontological characteristics of artworks and depicted entities. First, using IICONGRAPH~\cite{sartini2024iicongraph}, we categorized each depiction as either iconographic or pre-iconographic. Next, we examined whether the availability of images associated with the depicted entities influenced ranking quality. We also analyzed whether entities linked to artworks in the training graph (\textit{with artwork link}) exhibited different ranking behavior compared to those without such links (\textit{without artwork link}), under the condition that all associated artworks appeared only in the validation or test sets. Additionally, we investigated whether the input text explicitly mentioned the depicted entity. To this end, we employed a rule-based phrase-matching algorithm to identify English labels. For instance, in our motivating example shown in Fig.~\ref{fig:mot-example}, the entity ``ruins'' is not explicitly mentioned in the text. Finally, we assessed the impact of specific knowledge graph predicates by verifying whether the entity appeared as a subject in at least one triple containing a given predicate within the subgraph, thereby enabling us to evaluate the relevance of these predicates to the entity representations.

As shown in Fig.~\ref{fig:feat_importance}, \modelname{} significantly outperforms MIMIC in several cases, corroborating the findings of RQ1. Notably, substantial improvements are observed when the entity is not explicitly mentioned in the text. This suggests that contextual embeddings derived from the knowledge graph enhance the expressiveness of entity representations beyond what is available in the textual description alone. Moreover, performance gains are particularly pronounced for pre-iconographic entities, which are typically more abstract and thus benefit more from ontological relations such as \textit{subclass of}. Similarly, entities associated with predicates such as \textit{part of}, \textit{subclass of}, or \textit{instance of} tend to yield better results, as these relations provide valuable structural and semantic context.
Furthermore, entities linked to artworks in the graph (\textit{w/ artwork link}) exhibit improved rankings, confirming that modeling how an entity is visually depicted through entity–artwork links enriches its representation. Conversely, for entities without artwork links (\textit{w/o artwork link}), MIMIC outperforms \modelname{}, further confirming that integrating exemplifying information to depicted entities leads to improved rankings.

\begin{table*}[!t]
    \caption{Top-4 ranking comparison for ``\textit{Odysseus and Polyphemus}'' (\textbf{Q20537891}), a painting by Arnold Böcklin (1896), between \modelname{} and MIMIC~\cite{luo2023multi}. Ground truth entities are highlighted.}
    \label{tab:Q20537891_single_column}
    \centering
    \setlength{\tabcolsep}{6pt}
\begin{tabularx}{\textwidth}{cc>{\centering\arraybackslash}X>{\centering\arraybackslash}X}
    \toprule
    \textbf{Image \& Text} & & \textbf{MIMIC} & \textbf{\modelname{}} \\
    \toprule

    \multirow{2}{5cm}{\includegraphics[width=\linewidth]{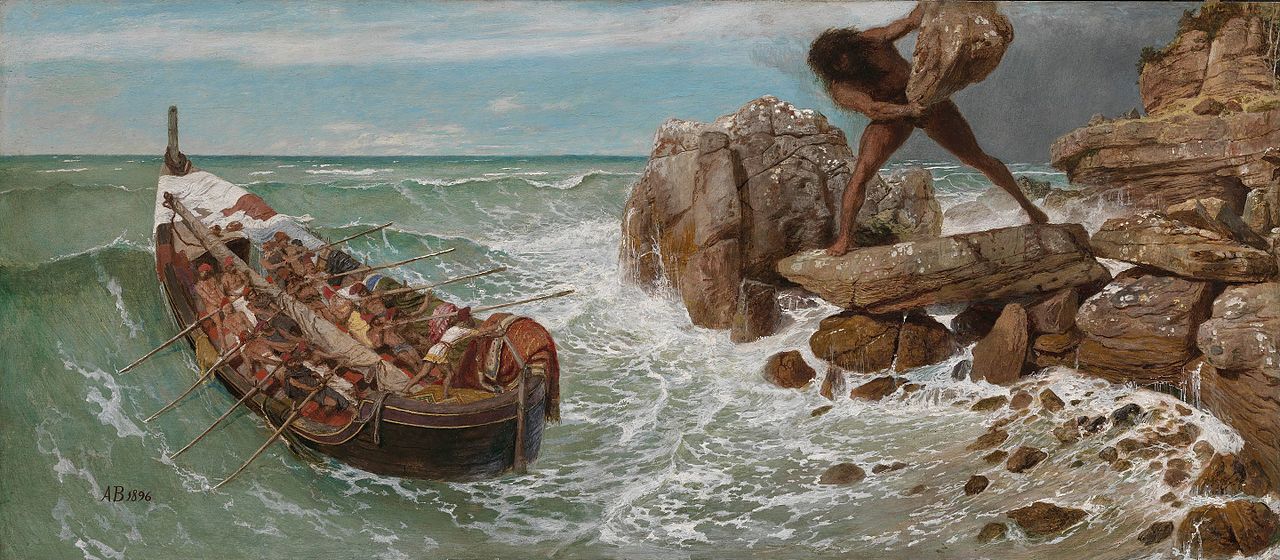}} & 
    \scriptsize{1} &
    \includecandidate{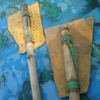}{\textbf{Oar.} water-borne propulsion} &
    \includecandidate[entcolor1]{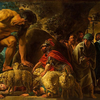}{\textbf{Polyphemus} giant son of Poseidon.}\\

    & 
    \scriptsize{2} & 
    \includecandidate[entcolor1]{Q193567}{\textbf{Polyphemus}. giant son of Poseidon.} &
    \includecandidate[entcolor2]{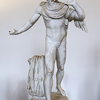}{\textbf{Odysseus}. Greek king of Ithaca.}\\

    \multirow{2}{5cm}{\scriptsize{\textit{The artwork ``Odysseus and Polyphemus'' represents a pivotal moment from Homer's epic poem, where Odysseus taunts the one-eyed Cyclops Polyphemus as he prepares to hurl a rock at the escaping boat of Odysseus and his crew.}}} & 
    \scriptsize{3} & 
    \includecandidate{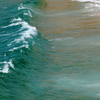}{\textbf{wind wave}. waves that occur on water} &
    \includecandidate{Q2003749}{\textbf{oar}. water-borne propulsion} \\

    & 
    \scriptsize{4} & 
    \includecandidate{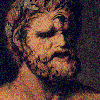}{\textbf{cyclops}. giant in Greek mythology.} &
    \includecandidate{Q128430}{\textbf{cyclops}. giant in Greek mythology.}\\\midrule

    \multicolumn{4}{l}{
    \textbf{Ground Truth:}
    \begin{tabularx}{0.7\linewidth}{X@{\hspace{12pt}}X@{\hspace{12pt}}}
        \includecandidate[entcolor1]{Q193567}{\textbf{Polyphemus}. giant son of Poseidon.} &
        \includecandidate[entcolor2]{Q47231}{\textbf{Odysseus}. Greek king of Ithaca.} \\
    \end{tabularx}
    }\\
    \bottomrule
\end{tabularx}
    
\end{table*}

Additionally, to qualitatively illustrate the differences in model behavior, we present an example from the test set in Table~\ref{tab:Q20537891_single_column}. In this case, \modelname{} correctly ranks both ground-truth entities—\textit{Polyphemus} and \textit{Odysseus}—at the top, while MIMIC ranks only one of them within the top-4. This example also shows that both models retrieve semantically plausible candidates; however, \modelname{}'s scores align more closely with the entities considered most significant according to Wikidata's annotations. Furthermore, \modelname{} presents a semantically cohesive ranking, suggesting that the KG entity encoding produces more semantically valuable information, thereby improving the model's expressiveness.

\subsubsection{Ablation Study}
To address RQ3, Table~\ref{tab:ablation} presents a unified ablation study that jointly evaluates variations in graph structure and node feature encoding, allowing us to assess the respective contributions of ontological context and input representations to entity linking performance.

Firstly, we evaluated the impactfulness of ontological information within \graphname{}, identifying two graph versions, namely the \textit{full} and the \textit{ablated} graph. The former encodes rich, heterogeneous contextual relations around each \textit{Depicted Object}, capturing how depicted entities are factually expressed and situated within broader semantic and ontological structures. These relations connect entities to types, attributes, and other contextual elements that shape their meaning beyond their immediate visual appearance. In contrast, the ablated graph retains only a minimal shared ontological link—(\textit{Depicted Object, instance of, Class})—removing most other contextual edges. In addition, we made a complementary ablation on raw node features encoding. Specifically, we compared three encoding strategies. In the textual setting, each node is described solely by a CLIP text embedding derived from its Wikidata label and description, prioritizing linguistic semantics. In the multimodal setting, text and image embeddings are combined via summation so that entities with available visuals can leverage both modalities (with zero-padding used when images are missing). Finally, the hybrid configuration uses visual features for artwork nodes while encoding all other entities textually, reflecting the intuition that artworks are best recognized visually whereas contextual entities (e.g., people, places, concepts) are better characterized through language.

Across all encoding strategies, the results in Table~\ref{tab:ablation} show a clear and consistent drop in performance when moving from the full graph to the ablated graph. This indicates that rich ontological and contextual relations play a crucial role in enabling more expressive and discriminative representations of depicted entities, helping the model capture how they are factually grounded in context rather than treated as isolated visual instances.

With respect to node feature encodings, the hybrid approach yields the best overall ranking performance in terms of MAP and MRR on the full graph, suggesting that a domain-aware combination of visual and textual signals provides the most balanced and effective representation. Interestingly, the purely textual model achieves the highest Recall@10, demonstrating that textual embeddings alone remain highly competitive, particularly in early retrieval settings. This is practically meaningful since end users typically inspect only the top-$k$ results, making Recall@10 a valuable indicator of real-world retrieval utility.

\begin{table*}[!t]
\centering
\caption{Ablation study on the effect of both contextual factors (i.e.,~Full Graph) and node feature encoding variations (i.e.,~Textual, Hybrid, and Multimodal). Numbers in \textbf{bold} indicate the best results.}
\resizebox{\textwidth}{!}{%
\begin{tabular}{clccc}
\toprule
Full Graph & Encoding   & R@10               & MAP                & MRR                \\ \midrule
\xmark     & Multimodal & 60.82 ($\pm$ 1.42) & 47.11 ($\pm$ 1.80) & 65.44 ($\pm$2.17)  \\
\xmark     & Hybrid     & 60.44 ($\pm$ 2.32) & 46.96 ($\pm$ 3.39) & 65.64 ($\pm$ 3.55) \\
\xmark     & Textual    & 61.42 ($\pm$ 1.92) & 47.87 ($\pm$ 2.72)  & 66.74($\pm$ 3.05)  \\
\cmark     & Multimodal & 62.05 ($\pm$ 0.38) & 48.28 ($\pm$ 0.83) & 67.53 ($\pm$ 1.09) \\
\cmark     & Hybrid     & 61.87 ($\pm$ 0.62) & \textbf{48.81 ($\pm$ 0.69)} & \textbf{67.93 ($\pm$ 1.04)} \\
\cmark     & Textual    & \textbf{62.31 ($\pm$ 0.57)} & 48.45 ($\pm$ 0.70) & 66.88 ($\pm$ 0.51) \\ \bottomrule
\end{tabular}}
\label{tab:ablation}
\end{table*}

\section{Conclusion} \label{sec:conclusion}

In this paper, we introduced \modelname{}, a context-aware framework for multimodal entity linking in the arts that integrates vision and language with structured relational knowledge. By combining a vision--language backbone with a heterogeneous graph encoder, \modelname{} leverages contextual dependencies to improve entity ranking and disambiguation in a challenging discovery setting, where relevant entities may be implicit and textual mention spans are not pre-defined. Experiments on the proposed \graphname{} dataset show consistent gains over strong baselines, highlighting the benefit of domain-relevant knowledge for robust multimodal linking.

A key strength of \modelname{} lies in graph-driven contextual reasoning: entities are represented not only by their local multimodal evidence but also by their positions within a heterogeneous relational structure. This enables the model to capture correlations and dependencies that are difficult to infer from vision or text alone, particularly for abstract or implicitly referenced entities. While our study focuses on cultural heritage, the proposed design provides a general blueprint for knowledge-aware discovery from multimodal and heterogeneous data.

Several directions for future work follow naturally. First, automatic and adaptive subgraph extraction could reduce reliance on manually curated predicates while improving scalability on large linked open data sources. Second, integrating explainable AI mechanisms—e.g., evidence attribution over graph neighborhoods and cross-modal interactions—could increase transparency and trustworthiness for downstream users. Finally, extending the framework to handle temporal or evolving knowledge graphs and to improve robustness under incomplete or noisy knowledge represents a promising avenue toward more reliable multimodal discovery systems.

\section*{Acknowledgments}
We acknowledge the CINECA award under the ISCRA initiative, which gave us access to computing resources and support. A Ph.D.~fellowship funds Raffaele Scaringi's research within the Italian ``D.M.~n.~352, April 9, 2022'' -- under the NRRP, Mission 4, Component 2, Investment 3.3 -- Ph.D.~project ``Automatic analysis of artistic heritage via Artificial Intelligence'', co-supported by Exprivia S.p.A. (CUP H91I22000410007). The HPI Information and Data Engineering Research School also partially funded this research.

\bibliographystyle{unsrt}  
\bibliography{references}  

\appendix

\section{Translation and summarization analysis}\label{sec:ap_translation_summarization}

\begin{figure}
    \centering
    \includegraphics[width=1\linewidth]{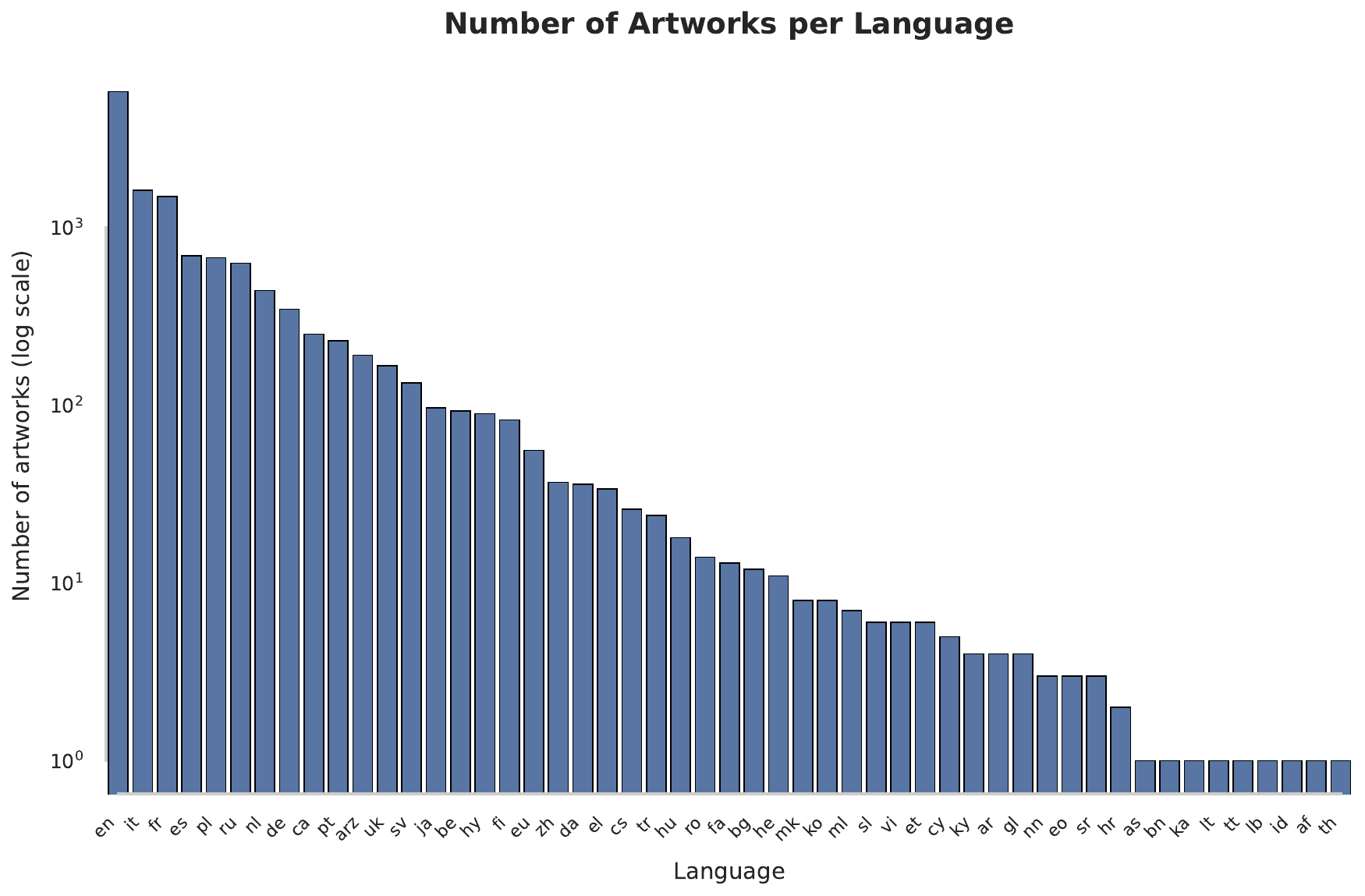}
    \caption{Distribution of artwork Wikipedia articles ($x$-axis) by original language ($y$-axis).}
    \label{fig:artworks-per-lang}
\end{figure}

We extracted Wikipedia articles from the snapshot updated at 2023/11/01\footnote{\url{https://huggingface.co/datasets/wikimedia/wikipedia}}, matching Wikipedia titles and languages from Wikidata. When the title was not found in the snapshot, we expanded the search using the redirect query \url{https://<LANG>.wikipedia.org/w/api.php?action=query&prop=redirects}, since some articles may have changed their titles during the time span between Wikidata's and Wikipedia's snapshots.
For each artwork that has no English Wikipedia article, we translated one article in a different language following these steps:
\begin{enumerate}
    \item Filter articles and keep the languages supported by NLLB
    \item Choose the longest article
    \item Split the article into paragraphs;
    \item Translate each paragraph independently;
    \item Concatenate the translated paragraphs.
\end{enumerate}
Figure~\ref{fig:artworks-per-lang} shows the proportion of languages from the selected articles in the dataset.

We balanced the length and level of detail in the Wikipedia articles summarizing them using Llama 3~\cite{grattafiori2024llama3herdmodels}. For each article, we used the following prompt to extract the summary:

\noindent
\framebox{\parbox{\dimexpr\linewidth-2\fboxsep-2\fboxrule}{\itshape%
`Summarize the following Wikipedia article related to an artwork in two sentences, with a maximum of 77 tokens. Go straight to the point, starting the response saying something like ``The artwork \texttt{<artwork\_name>} represents...''. Focus on the visual aspects of the article, including all the depicted entities. Get the info from the following article: \texttt{\{wikiarticle\}}'}}

Examples:

Original article:

\noindent
\framebox{\parbox{\dimexpr\linewidth-2\fboxsep-2\fboxrule}{\itshape%
Triumph of Chastity is an oil-on-canvas allegorical painting by the Italian Renaissance artist Lorenzo Lotto, created c.\ 1530, now in the Rospigliosi Pallavicini collection in Rome. It is signed at bottom right ``Laurentius Lotus'' and shows a female personification of Chastity (left) driving away Cupid and Venus. The Venus is based on that on an ancient Roman sarcophagus now in the Vatican Museums.
}}

Summary:

\noindent
\framebox{\parbox{\dimexpr\linewidth-2\fboxsep-2\fboxrule}{\itshape%
The artwork ``Triumph of Chastity'' represents a powerful allegory, featuring a personification of Chastity driving away Cupid and Venus. The painting depicts Chastity, flanked by Cupid and the goddess Venus, based on an ancient Roman sarcophagus now housed in the Vatican Museums.
}}

\begin{figure}[!t]
    \centering
    \includegraphics[width=1\linewidth]{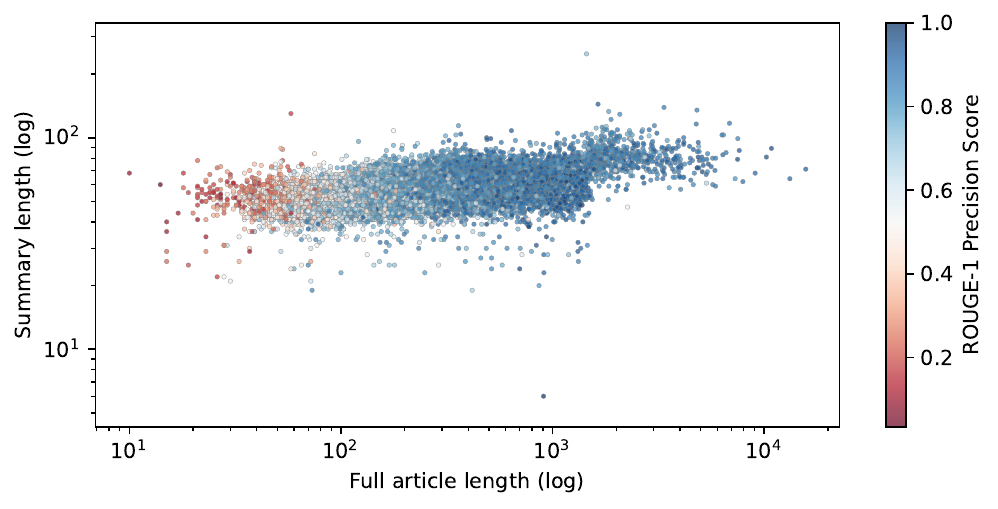}
    \caption{Length in words of the original articles and their summaries. The color represents the ROUGE-1 precision for each article-summary pair.}
    \label{fig:article_vs_summary}
\end{figure}

\begin{figure}[!t]
    \centering
    \includegraphics[width=1\linewidth]{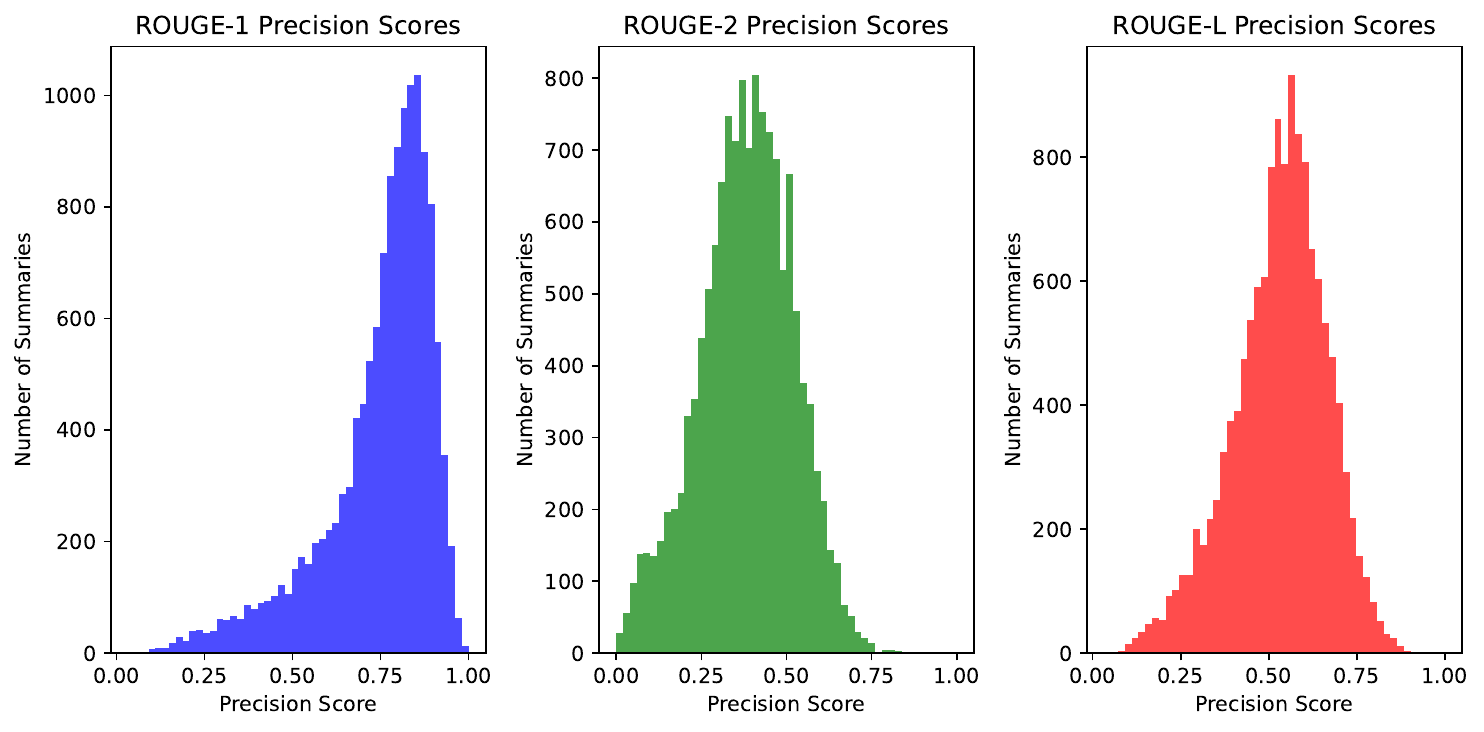}
    \caption{Distribution of ROUGE precision between the summaries and the full articles.}
    \label{fig:rouge_distributions}
\end{figure}

The result is a more balanced article length, which can be seen in Figure~\ref{fig:article_vs_summary}. We see that the summary length is concentrated between 30 and 100 word texts, while the original articles ranged between 20 and more than 1000 words.
We measure the quality of the summaries by measuring ROUGE precision scores \cite{lin-2004-rouge}. In Figure~\ref{fig:article_vs_summary}, we observe that as expected lower precision is present in the shorter original articles, but the majority of article summaries have a large overlap with the original articles. This can be further observed in Figure~\ref{fig:rouge_distributions} and Table~\ref{tab:rouge}, where the ROUGE-1 distributions is clearly concentrated in the higher values. Nevertheless, we observe a drop in ROUGE-2 precision, likely due to language simplification.

\begin{table*}[!t]
\centering
\caption{ROUGE precision metrics}
\begin{tabular}{lrrr}
\toprule
 & Rouge1 & Rouge2 & RougeL \\
\midrule
mean & 0.745735 & 0.381481 & 0.530465 \\
std & 0.158000 & 0.138832 & 0.134143 \\
\bottomrule
\end{tabular}
\label{tab:rouge}
\end{table*}

\end{document}